\documentclass[sigconf]{acmart}
\usepackage{multirow} 
\usepackage{array}   
\usepackage{tabularx}      
\usepackage{booktabs}       
\usepackage{ragged2e}       
\usepackage{siunitx} 
\newcolumntype{Y}{>{\RaggedRight\arraybackslash}X} 
\usepackage{makecell}
\usepackage{graphicx} 
\usepackage{float} 
\usepackage{subfigure} 
\usepackage{caption}
\usepackage{subcaption}
\usepackage{tcolorbox}  
\usepackage{xcolor}     
\usepackage{stfloats} 
\usepackage{hyperref}
\usepackage{enumitem}
\setlist[itemize]{leftmargin=*, nosep} 
\usepackage{threeparttable}
\usepackage{tikz}
\makeatletter
\DeclareRobustCommand{\circled}[1]{\tikz[baseline=(char.base)]{
    \node[
        shape=circle, draw=black, line width=0.6pt,
        inner sep=0pt, minimum size=1.0em,
        font=\normalfont\small,
        outer sep=0pt
    ] (char) {#1};}}
\makeatother
\usepackage{algorithm}
\usepackage{algorithmic}

\AtBeginDocument{%
  }

\setcopyright{acmlicensed}
\copyrightyear{2025}
\acmYear{2025}
\acmDOI{XXXXXXX.XXXXXXX}
\acmConference[MM '25]{Proceedings of the 33rd ACM International Conference on Multimedia}{October 27--31, 2025}{Dublin, Ireland}
\acmBooktitle{Proceedings of the 33rd ACM International Conference on Multimedia (MM '25), October 27-31, 2025, Dublin, Ireland}
\acmISBN{978-1-4503-XXXX-X/2025/10}

\begin{document}

\title{Memory-Augmented Dual-Decoder Networks for Multi-Class Unsupervised Anomaly Detection}

\author{Jingyu Xing}
\affiliation{%
  \institution{Sichuan University}
  \city{Chengdu}
  \country{China}
}
\email{xingjingyuok@stu.scu.edu.cn}

\author{Chenwei Tang}
\affiliation{%
  \institution{Sichuan University}
  \city{Chengdu}
  \country{China}
}
\email{tangchenwei@scu.edu.cn}

\author{Tao Wang}
\affiliation{%
  \institution{Sichuan University}
  \city{Chengdu}
  \country{China}
}
\email{twangnh@gmail.com}

\author{Rong Xiao}
\affiliation{%
  \institution{Sichuan University}
  \city{Chengdu}
  \country{China}
}
\email{rxiao@scu.edu.cn}

\author{Wei Ju}
\affiliation{%
  \institution{Sichuan University}
  \city{Chengdu}
  \country{China}
}
\email{juwei@scu.edu.cn}

\author{Ji-Zhe Zhou}
\affiliation{%
  \institution{Sichuan University}
  \city{Chengdu}
  \country{China}
}
\email{jzzhou@scu.edu.cn}

\author{Liangli Zhen}
\affiliation{%
  \institution{Institute of High Performance Computing, A*STAR}
  \country{Singapore}
}
\email{mi.lianglizhen@gmail.com}

\author{Jiancheng Lv}
\affiliation{%
  \institution{Sichuan University}
  \city{Chengdu}
  \country{China}
}
\email{lvjiancheng@scu.edu.cn}

\renewcommand{\shortauthors}{Trovato et al.}

\begin{abstract}
Recent advances in unsupervised anomaly detection (UAD) have shifted from single-class to more practical multi-class scenarios. In such complex contexts, the increasing pattern diversity has brought two critical challenges to reconstruction-based approaches: (1) \textbf{over-generalization} — anomalies that are subtle or share compositional similarities with normal patterns may be reconstructed with unexpectedly high fidelity, making them difficult to distinguish from normal instances; and (2) \textbf{insufficient normality reconstruction} — complex normal features, such as intricate textures or fine-grained structures, may not be faithfully reconstructed due to the model's limited representational capacity, resulting in false positive detections. Existing methods typically focus on addressing the former, which unintentionally exacerbate the latter, resulting in inadequate representation of intricate normal patterns. To concurrently address these two challenges, we propose a novel \textbf{M}emory-augmented \textbf{D}ual-\textbf{D}ecoder \textbf{Net}works (\textbf{MDD-Net}). This network includes two critical components: a Dual-Decoder Reverse Distillation Network (DRD-Net) and a Class-aware Memory Module (CMM). Specifically, the DRD-Net incorporates a restoration decoder designed to recover normal features from synthetic abnormal inputs and an identity decoder to reconstruct features that maintain the anomalous characteristics. By exploiting the discrepancy between features produced by these two decoders, our approach refines anomaly scores beyond the conventional encoder-decoder comparison paradigm, effectively reducing false positives and enhancing localization accuracy, particularly for complex normal instances. Furthermore, the CMM explicitly encodes and preserves class-aware normal prototypes, actively steering the network away from anomaly reconstruction. Comprehensive experimental results across several benchmarks demonstrate the superior performance of our MDD-Net framework over current state-of-the-art approaches in multi-class UAD tasks.
\end{abstract}



\maketitle
\begin{figure}[ht] 
  \centering
  \includegraphics[width=0.9\columnwidth]{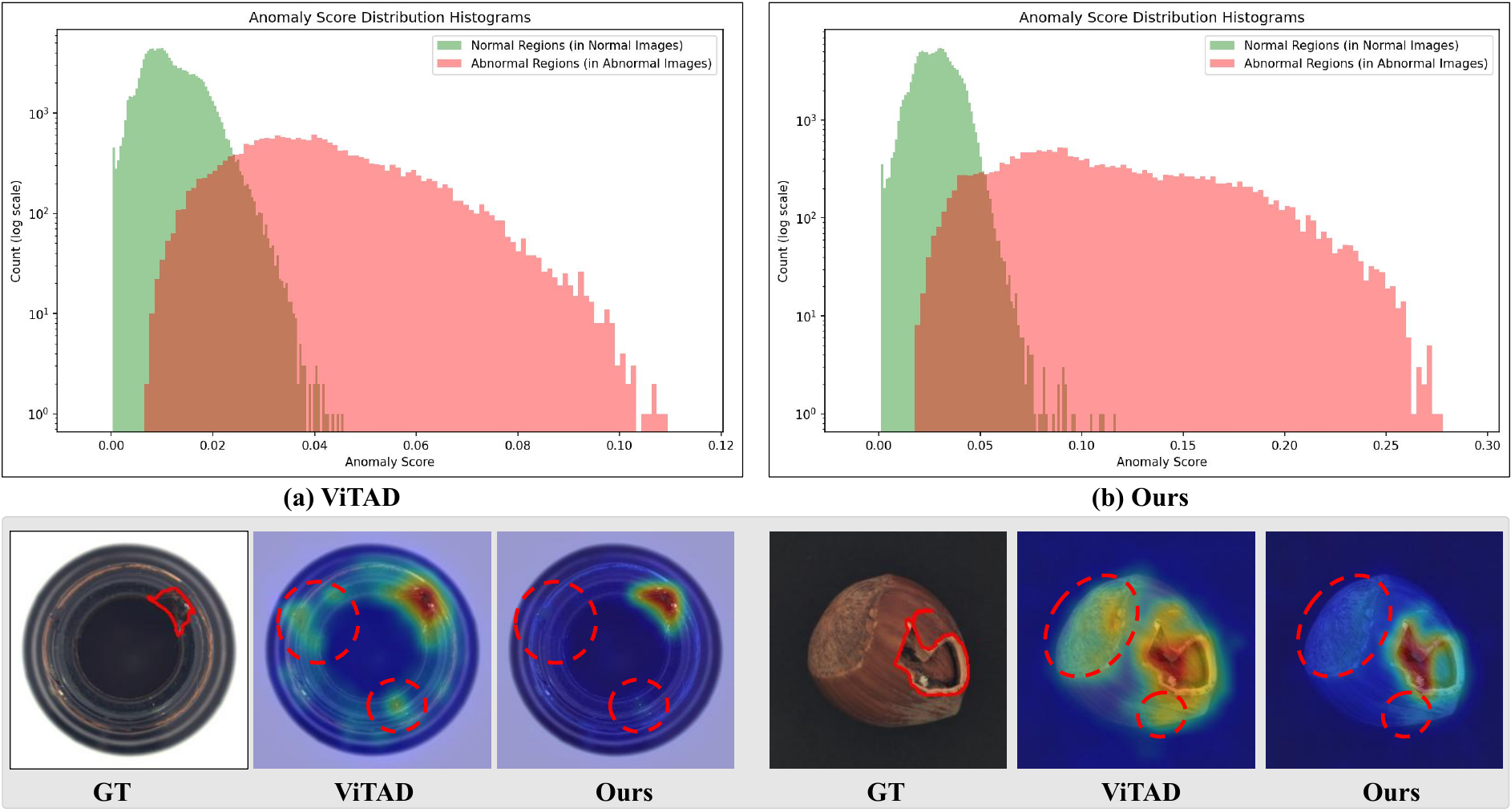} 
  \caption{Reconstruction error histograms and detection heatmap results of ViTAD~\cite{vitad} \& Ours on MVTec AD~\cite{mvtec}.}
  \Description[histograms]{histograms}
  \label{fig:hist}
\end{figure}
\section{Introduction}

Unsupervised Anomaly Detection (UAD)~\cite{patchcore,cflow,nsa} aims to detect and locate anomalous regions in images using only normal samples to train the model, showing application potential in industrial quality inspection~\cite{industry,mvtec,nuclear}, medical disease screening~\cite{medical1,medical2,medical3}, and video surveillance~\cite{video1,video2,video3}. Previous UAD methods follow the single-class setting~\cite{padim,memkd,simplenet,realnet,diffusionad} by training separate models for different classes, which is resource-intensive and impractical for real-world deployment. Recent advancements have shifted toward a more practical yet challenging multi-class UAD paradigm~\cite{uniad,omnial}, which employs a unified model to detect anomalies across diverse categories without requiring category-specific fine-tuning. 

Most existing multi-class UAD methods rely on feature reconstruction~\cite{rd,rdpp,kdsdd,uniad} to model the distribution of normal data. The underlying assumption is that, when trained solely on normal samples, the model will consistently reconstruct normal features, while producing large reconstruction errors for anomalous inputs. Architecturally, these approaches typically adopt a encoder-decoder framework: (1) a pre-trained encoder (Teacher) extracts discriminative multiscale feature representations, and (2) a decoder (Student) takes the teacher's output embeddings as input and learns to reconstruct these features by minimizing a reconstruction loss computed using normal training data. During inference, anomalies are detected by evaluating the discrepancy between encoder's original input features and the corresponding reconstructions produced by the decoder.

Subsequent improvements focus on amplifying the reconstruction discrepancy of anomalous and normal inputs through degraded attention~\cite{uniad}, memory mechanism~\cite{memkd,rlr}, and vector quantization~\cite{hvq-trans}. These methods aim to systematically constrain the decoder's representation capability for anomalous features. However, these strategies unintentionally exacerbate the  \textbf{insufficient normality reconstruction} issue, which arises when certain normal features are reconstructed with unexpectedly high errors, causing elevated false positive rates and degraded anomaly localization. As illustrated in Figure~\ref{fig:hist}, the reconstruction error histogram of ViTAD~\cite{vitad} indicates that certain normal regions incur unexpectedly large errors. Correspondingly, its localization results reveal high edge uncertainty and substantial false positives within normal areas. We attribute this to the decoder's limited representational capacity, which hinders accurate reconstruction of intricate normal features. Furthermore, the foundational assumption of reconstruction-based approaches—that abnormal inputs yield significantly higher reconstruction errors—does not universally hold. In particular, certain anomalous regions can be reconstructed with high fidelity, resulting in the well-recognized problem of \textbf{over-generalization}~\cite{overgeneralization1,overgeneralization2,rd}. This typically arises when anomalies are subtle or share compositional similarities with normal training patterns, enabling the decoder to generalize and inadvertently reconstruct them. Paradoxically, this reveals a critical observation: the decoder may succeed in reconstructing subtle or shared-pattern anomalies, while simultaneously failing to reconstruct complex normal features.

To overcome the aforementioned challenges, we propose a \textbf{M}emory augmented \textbf{D}ual-\textbf{D}ecoder \textbf{Net}works (\textbf{MDD-Net}) framework that mitigates both the \textbf{insufficient normality reconstruction} for hard-normal features and the \textbf{over-generalization} of anomalous patterns. MDD-Net comprises two key components: a Dual-Decoder Reverse Distillation Network (DRD-Net) and a Class-aware Memory Module (CMM). Specifically, DRD-Net includes two decoders: a restoration decoder that reconstructs normal feature representations and an identity decoder that retains semantic information specific to anomalies. By leveraging decoupled features from these decoders along with the teacher encoder’s features, we generate anomaly maps that reduce false positives and improve localization accuracy, particularly for complex normal patterns. Meanwhile, CMM is designed to learn and store class-aware normal prototypes. By constraining the restoration decoder to align with these prototypes, CMM suppresses the reconstruction of anomalous patterns and promotes larger reconstruction errors for anomalies, thereby alleviating the over-generalization issue.

Our main contributions are three-fold:
\begin{itemize}
    \item We propose MDD-Net, a unified framework that simultaneously addresses the insufficient reconstruction of hard-normal features and the over-generalization of subtle anomalies. Extensive experiments demonstrate its superiority, achieving an average improvement of 3.6\% over previous state-of-the-art methods on the MVTec AD benchmark.
    \item We introduce DRD-Net, a dual-decoder architecture that decouples reconstruction objectives. By comparing the anomaly maps derived from both decoders and the teacher encoder, DRD-Net effectively reduces false positives and enhances localization precision.
    \item We design CMM, a class-aware memory mechanism that learns normal prototypes and actively prevents the decoder from reconstructing anomalous content. This amplifies feature discrepancies in abnormal regions and mitigates over-generalization.
\end{itemize}

\section{Related Work}
Most existing unsupervised anomaly detection (UAD) methods operate in a single-class setting, where separate models are trained for each object category. These approaches can be divided into three categories: 1) Synthesizing-based methods typically generate pseudo-anomalies by synthesizing anomalous regions~\cite{cutpaste,nsa} or perturbing normal features~\cite{simplenet,glass}, frequently serving as plug-and-play modules to enhance other frameworks. 2) Embedding-based methods utilize pretrained networks to extract feature embeddings, then employ density estimation~\cite{cflow,msflow} or memory banks~\cite{patchcore,padim} to model normality distribution.  3) Reconstruction-based methods rely on the hypothesis that models trained on normal samples reconstruct normal regions effectively but fail on anomalies, which is further subdivided into image reconstruction~\cite{draem,ocr-gan} and feature reconstruction~\cite{rd,destseg}. While effective in constrained environments, such approaches are not scalable for real-world applications that involve diverse categories and frequent deployment demands. 

To overcome this limitation, recent research has shifted toward multi-class UAD (MUAD), which enables a unified model to detect anomalies across multiple categories. However, due to the diversity inherent in multi-class normal patterns, traditional Synthesizing- and Embedding-based methods struggle to model normality distribution effectively. And benefiting from lower computational overhead, feature reconstruction approaches have become demonstrably competitive in this context. A representative example is UniAD~\cite{uniad}, which introduces a standard feature reconstruction framework that eliminates the need for category-specific training. Building upon this direction, several other MUAD methods have been proposed, including DiAD~\cite{diad}, HVQ-Trans~\cite{hvq-trans}, and MambaAD~\cite{mambaad}. One of the primary challenges in MUAD is the so-called identical shortcuts failure, where both normal and abnormal samples are reconstructed with high fidelity, thereby blurring the distinction between the two. This issue is attributed to the high diversity of normal patterns in multi-class settings, which inadvertently drives the model to generalize even to unseen, anomalous patterns~\cite{uniad}. To address this, various strategies have been explored. For instance, UniAD~\cite{uniad} introduces a neighbor-masked attention module and a feature-jitter strategy to mitigate these shortcuts. HVQ-Trans~\cite{hvq-trans} proposes a vector quantization (VQ) Transformer model that induces large feature discrepancies for anomalies. DiAD~\cite{diad} employs diffusion models to address multi-class UAD settings. ViTAD~\cite{vitad} abstracts a unified feature-reconstruction UAD framework and employs Transformer building blocks. MambaAD~\cite{mambaad} explores the recently proposed State Space Model (SSM) in the context of multi-class UAD. 

Despite these advances, existing methods largely focus on improving anomaly separability through architectural innovations or attention mechanisms, often overlooking the dual challenges of over-generalization and the insufficient reconstruction of complex normal features. Our proposed MDD-Net addresses this gap by explicitly tackling both limitations within a unified framework.

\section{Our Proposed Method}

\begin{figure*}[t]
\centering
\includegraphics[width=\textwidth]{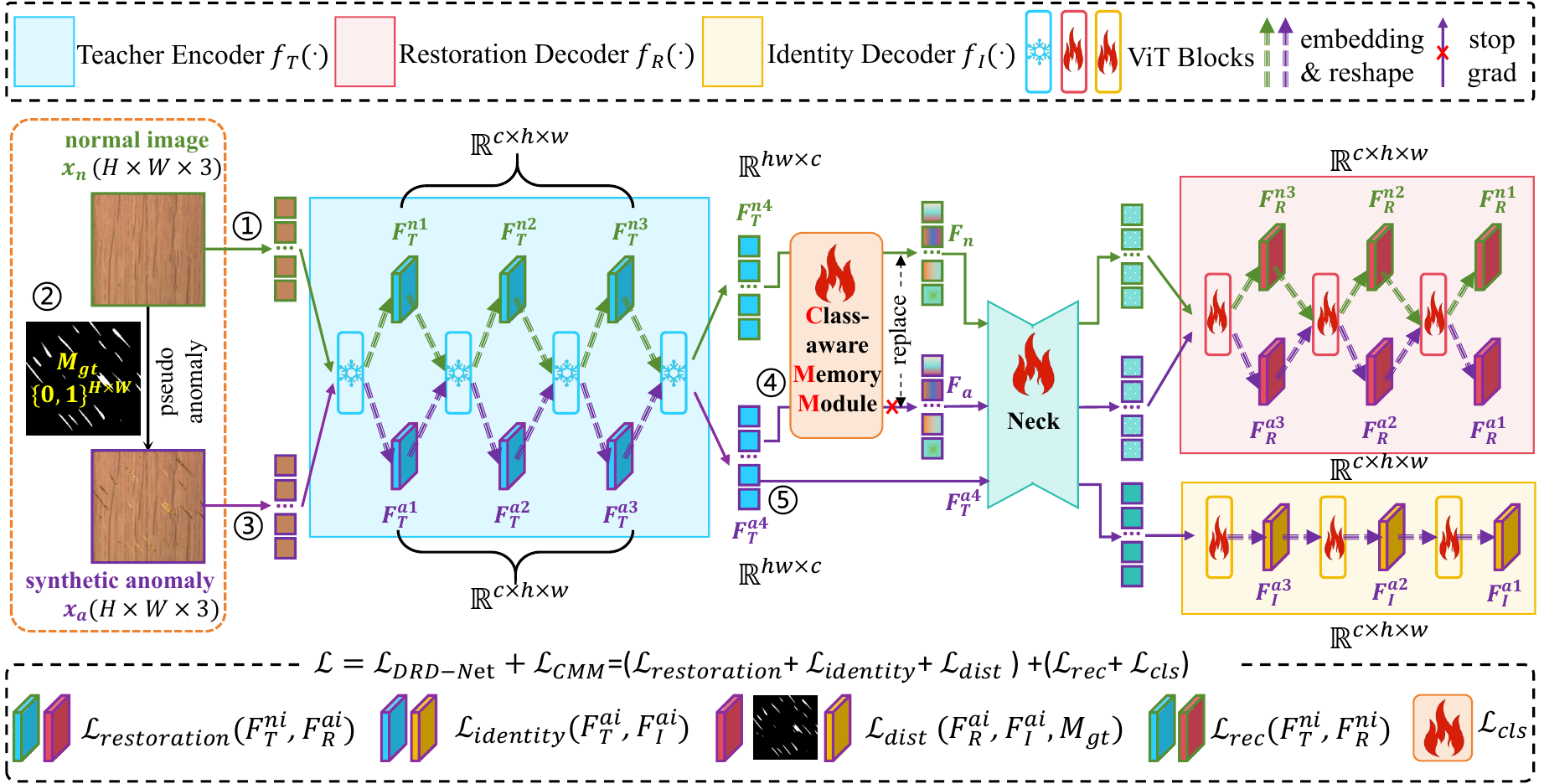}
\caption{Training Process of MDD-Net, which is divided into five steps: \circled{1} \underline{Normality Reconstruction}: Normal images are input and reconstructed through the frozen Teacher Encoder, CMM, Neck, and Restoration Decoder to update CMM parameters with $\mathcal{L}_{rec}$ and $\mathcal{L}_{cls}$ loss. \circled{2}  \underline{Anomaly Synthesis}: Pseudo-anomaly samples containing anomalous images and corresponding masks are generated using an image-level anomaly synthesis method. \circled{3} \underline{Anomalous Feature Extraction}: The synthesized anomalies are fed into the frozen Teacher Encoder to extract features containing anomalous information. \circled{4} \underline{Anomaly Restoration}: The anomalous features extracted in step \circled{3} are processed sequentially through CMM, Neck, and Restoration Decoder for anomalous feature elimination, forming the restoration loss term $\mathcal{L}_{restoration}$. \circled{5} \underline{Anomaly Reconstruction}: The anomalous features extracted in step \circled{3} are reconstructed through Neck and Identity Decoder, constituting the identity loss term $\mathcal{L}_{identity}$. Additionally, to enhance feature discrepancies between the two decoders in anomalous regions, an anomaly mask supervision mechanism is introduced, resulting in the discrepancy loss term $\mathcal{L}_{dist}$.}
\Description[MDD-Net]{MDD-Net}
\label{net}
\end{figure*}

\subsection{Overview}
As illustrated in Figure \ref{net}, the proposed MDD-Net comprises four main components: a frozen teacher encoder, a learnable neck module, a class-aware memory module (CMM), and two decoders (a Restoration Decoder to recover normal features from anomalous inputs and an Identity Decoder to preserve anomaly semantics). The entire architecture is constructed using ViT blocks and MLP layers. Specifically, the teacher encoder consists of 12-layer ViT blocks initialized with pre-trained DINO-S~\cite{dino} weights, organized into four stages with three ViT blocks each. Both decoders utilize randomly initialized ViT structures reduced to three stages (9 layers in total, 3 layers per stage), while the neck module implements a lightweight single-layer linear transformation to minimize computational overhead. During training, we generate synthetic anomalous samples by following DRAEM~\cite{draem} to create paired normal-abnormal images. These synthetic anomalies are processed through the teacher encoder and subsequently routed to either Restoration Decoder or Identity Decoder. During inference, the anomaly map is computed by collectively analyzing feature representation discrepancies of the teacher encoder, Restoration Decoder and Identity Decoder. The algorithm for MDD-Net's training and inference is provided in the Supplementary Materials Section A.3 .

\subsection{Dual-Decoder Reverse Distillation Network}
To address the issues of false positives and localization inaccuracies caused by insufficient reconstruction of hard-normal patterns, we propose a novel framework incorporating pseudo anomalies and a Dual-Decoder Reverse Distillation Network (DRD-Net). It consists of a Restoration Decoder optimized to minimize the discrepancy between reconstructed features of anomalies and the teacher encoder's features of normal counterparts, and an Identity Decoder designed to preserve the structural fidelity between reconstructed features of anomalies and its own features obtained from the teacher encoder. We argue that both decoders possess identical structures and comparable parameter capacities, so their representational capacities remain inherently correlated. Therefore, if one decoder exhibits sub-optimal reconstruction performance of hard-normal patterns, the other would inevitably demonstrate similar deficiencies. By comparing output differences between them, we can effectively mitigate the challenges stemming from such reconstruction inadequacy. The training procedure unfolds as follows:

\textbf{1. Synthetic Anomaly Generation.} First, we follow DRAEM~\cite{draem} for anomaly synthesis. Given a normal image $x_n \in \mathbb{X} \subseteq\mathbb{R}^{H \times W \times 3}$, we generate its perturbed counterpart $x_a = \mathcal{A}(x_n)$ through an augmentation function $\mathcal{A}(\cdot)$ that blends Perlin noise with the Describable Textures Dataset (DTD)~\cite{dtd}. This process yields a synthetic anomaly image paired with its corresponding binary anomaly mask $M_{gt} \in \{0,1\}^{H \times W}$, which explicitly indicates perturbed regions.

\textbf{2. Hierarchical Feature Extraction.} Subsequently, the frozen teacher encoder $f_T(\cdot)$ processes both $x_n$ and $x_a$ to extract multi-stage features: 1) Spatial features ($i=1,2,3$): These features are reshaped from token sequences into 2D spatial layouts, forming hierarchical representations $F_T^{ni}, F_T^{ai} \in \mathbb{R}^{bs \times c \times h \times w}$ at resolutions $(h, w)$. 2) Neck features ($i=4$): The deepest layer outputs token sequences $F_T^{n4}, F_T^{a4} \in \mathbb{R}^{bs \times T \times c}$ where $T=hw$ denotes the number of tokens. The normal features $\{F_T^{ni}\}_{i=1}^3$ serve as reference targets, while $\{F_T^{ai}\}_{i=1}^4$ encode anomaly-perturbed semantics.

\textbf{3. Normal Prototype Replacement.} The CMM $\mathcal{M}(\cdot)$ (training details in Section \ref{section:CMM}) stores learnable normal prototypes. To suppress anomaly semantics in $F_T^{a4}$, CMM performs prototype replacement:
\begin{equation}
    F_{a} = \mathcal{M}(F_T^{a4}),
\end{equation}
where each token is substituted by linear combinations of its nearest normal prototypes from the memory bank. To prevent anomaly contamination during memory updating, gradient truncation is applied to features of synthetic anomalies after CMM retrieval. 

\textbf{4. Dual-Decoder Reverse Distillation.} Ultimately, Our framework employs two complementary decoders: 1) Restoration Decoder $f_R(\cdot): F_{a} \mapsto \left\{ F_R^{ai} \right\}_{i=1}^3$: The Restoration Decoder decodes $F_a$ to reconstruct spatial features of normal counterparts $\{F_T^{ni}\}_{i=1}^3$ , functioning like denoising. The restoration loss enforces feature alignment with the teacher encoder's normal features:
\begin{equation}
    \mathcal{L}_{restoration} = \sum_{i=1}^3 \left[1 - \frac{\langle F_R^{ai}, F_T^{ni} \rangle}{\|F_R^{ai}\| \|F_T^{ni}\|}\right].
\end{equation}
2) Identity Decoder $f_I(\cdot): F_T^{a4} \mapsto \left\{ F_I^{ai} \right\}_{i=1}^3$: To preserve anomaly semantics, the Identity Decoder directly decodes $F_T^{a4}$ to mimic $\{F_T^{ai}\}_{i=1}^3$. The identity loss maintains feature fidelity:
\begin{equation}
    \mathcal{L}_{identity} = \sum_{i=1}^3 \left[1 - \frac{\langle F_I^{ai}, F_T^{ai} \rangle}{\|F_I^{ai}\| \|F_T^{ai}\|}\right].
\end{equation}

\textbf{5. Discrepancy Amplification Learning.} Moreover, intuitively, \textbf{in normal regions} (where $M_{gt}^{ai}=0$), both the Restoration Decoder and Identity Decoder imitate the normal features generated by the teacher encoder, resulting in aligned feature representations with low distance. Conversely, \textbf{in anomalous regions} (where $M_{gt}^{ai}=1$), the Restoration Decoder endeavors to recover normal features while the Identity Decoder preserves anomalous semantics, thereby maximizing their divergence. This phenomenon exactly coincides with numerical distributions of the anomaly mask $M_{gt}^{ai}$. Therefore, to further amplify feature discrepancies in anomalous regions, we compute pixel-wise cosine distance maps $M_{RI}^{ai}(h, w)$ between two decoders. And the anomaly mask $M_{gt}^{ai}$ (downsampled to $h \times w$ resolution) supervises this divergence through an L1-loss:

\begin{equation}
    M_{RI}^{ai}(h, w) = 1 - \frac{\langle F_R^{ai}(h, w), F_I^{ai}(h, w) \rangle}{\|F_R^{ai}(h, w)\| \|F_I^{ai}(h, w)\|},
\end{equation}

\begin{equation}
    \mathcal{L}_{dist} = \sum_{i=1}^3  \| M_{RI}^{ai} - M_{gt}^{ai} \|_1.
\end{equation}

\textbf{6. Unified Training Objective of DRD-Net.} The complete optimization objective of DRD-Net combines all components:
\begin{equation}
    \mathcal{L}_{DRD-Net} = \mathcal{L}_{restoration} + \mathcal{L}_{identity} + \mathcal{L}_{dist}.
\end{equation}

During inference, input images are passed through the teacher encoder, Restoration Decoder and Identity Decoder to extract features at different levels. Both the Restoration-Identity Discrepancy (RID) and Teacher-Restoration Discrepancy (TRD) exhibit significant divergences between normal and anomalous images, providing a strong discriminative basis for anomaly detection (details in Section \ref{sec:asmsf}). 





\subsection{Class-aware Memory Module}
\label{section:CMM}
Following MemAE~\cite{memae}, the proposed CMM inherits: a memory bank storing normality prototypes and an attention-based sparse retrieval mechanism. Moreover, to address Multi-class UAD challenges, we integrate class information as supervision signals, enhancing the memory's class-discriminative capability. 


\textbf{1. Memory-based Representation.} The memory bank $\mathbf{M} \in \mathbb{R}^{N \times c}$ stores $N$ normal prototype vectors $\{m_i\}_{i=1}^N$ with dimension $c$. Given an encoded query $F \in \mathbb{R}^{T \times c}$, the module reconstructs $\hat{F}$ through weighted retrieval:
\begin{equation}
\widehat{F} = w\mathbf{M} = \mathop{\sum }\limits_{{i = 1}}^{N}{w}_{i}{m}_{i},
\end{equation}
where $w$ is a row vector with non-negative entries that sum to one and ${w}_{i}$ denotes the $i$-th entry of $w$. The hyper-parameter $N$ defines the maximum capacity of the memory. The hyper-parameter analysis is presented in the Supplementary Materials Section B.1 .

\textbf{2. Attention-based Sparse Retrieval.} The retrieval weights $w$ are computed via cosine similarities between feature $F$ and memory prototypes, following MemAE~\cite{memae}. To prevent anomaly reconstructions through dense prototype combinations, we apply sparse shrinkage on $w$ using a hard threshold $\lambda$ (typically $\lambda \in [1/N, 3/N]$), implemented via a differentiable ReLU approximation. 


\textbf{3. Normality Prototypes Learning.} We update the CMM to encode class-aware normal prototypes through reconstruction loss $\mathcal{L}_{rec}$ and classification loss $\mathcal{L}_{cls}$. First, to update CMM with typical normal prototypes, we introduce an auxiliary data flow of normal images passing through the Restoration Decoder for standard reconstruction. For normal image $x_n$, we compute:
\begin{equation}
    x_n \xrightarrow{f_T(\cdot)}F_T^{n4} \xrightarrow{\mathcal{M(\cdot)}} F_n \xrightarrow{f_R(\cdot)} \left\{ F_R^{ni} \right\}_{i=1}^3,
\end{equation}
\begin{equation}
    \mathcal{L}_{rec} = \sum_{i=1}^3 \left[1 - \frac{\langle F_R^{ni}, F_T^{ni} \rangle}{\|F_R^{ni}\| \|F_T^{ni}\|}\right].
\end{equation}

Then, to endow the normal prototypes with better class discrimination, we propose a novel classification mechanism that propagates global category priors to individual memory items. The detailed steps are as follows (the corresponding workflow is visualized in Supplementary Materials Section A.1 ): 

1) Assign Learnable Class Probability for Memory Items. Each memory item $m_i \in \mathbb{R}^c$ is assigned a learnable class probability vector $p_i \in \mathbb{R}^D$. These vectors $\{p_i\}_{i=1}^N$ implicitly capture the categorical attributes of prototypes, allowing them to specialize in class-specific or cross-class patterns during training. 

2) Aggregate Token-Level Class Probability. For an input sequence $F_T^{n4} \in \mathbb{R}^{T \times c}$ , the memory module reconstructs each token ${F_T^{n4}}^{(t)} \in \mathbb{R}^c$ at position $t \in \{1,\dots,T\}$ as : $F_{n}^{(t)} = \sum_{i=1}^N w_i^{(t)} m_i$, where $w_i^{(t)}$ denotes the attention weight between the $t$-th token and the $i$-th memory item. And the class prediction $\hat{y}^{(t)}$ for the $t$-th token is obtained by weighted aggregation:
\begin{equation}
\hat{y}^{(t)} = \sum_{i=1}^N w_i^{(t)} p_i.
\end{equation}

This formulation dynamically associates each token with a weighted combination of memory items' class semantics. 

3) Dense Supervision via Global Labels. Given the input image's class label $y$, we enforce consistency between all token-level predictions $\{\hat{y}^{(t)}\}_{t=1}^T$ and $y$ using a cross-entropy loss:
\begin{equation}
    \mathcal{L}_{cls} = \frac{1}{T} \sum_{t=1}^T  \left[ -\hat{y}^{(t)}_{y} + \log\left( \sum_{D=1}^D \exp({\hat{y}^{(t)}_c}) \right) \right], 
\end{equation}
where $\hat{y}^{(t)}_{y}$ is the predicted class probability corresponding to the true class $y$. 


The proposed classification mechanism enables the CMM to dynamically disentangle category-specific and cross-class shared prototypes: class-activated prototypes sharpen categorical distributions through gradient alignment with global labels, while shared prototypes maintain non-peaked probability profiles to capture multi-class patterns. 

Overall, the total objective of \textit{Normality Prototypes Learning} is:
\begin{equation}
\mathcal{L}_{CMM} = \mathcal{L}_{rec} + \mathcal{L}_{cls}.
\end{equation}

The $\mathcal{L}_{cls}$ synergizes with the $\mathcal{L}_{rec}$, jointly optimizing the CMM to compactly encode both normality and categorical semantics. The design requires no additional category annotations beyond standard classification labels, maintaining minimal complexity.

\subsection{Training objectives}
In summary, our framework involves two loss functions: the optimization loss for the DRD-Net and the updating loss for the CMM.
\begin{equation}
\mathcal{L} =    \mathcal{L}_{DRD-Net} + \mathcal{L}_{CMM}.
\end{equation}

Notably, our method demonstrates remarkable effectiveness through direct summation of these loss terms, eschewing complex weight-balancing mechanisms while achieving SoTA performance. 

\subsection{Anomaly Scoring via Multi-Source Fusion}
\label{sec:asmsf}
During inference, MDD-Net leverages two complementary reconstruction discrepancies RID and TRD to enhance anomaly detection. The detailed inferring process is available in Supplementary Materials Section A.2 .


\textbf{1. Restoration-Identity Discrepancy.} Since the Restoration Decoder $f_R$ recovers normal features from anomaly inputs using \textit{normal-replaced features} after the CMM, while the Identity Decoder $f_I$ preserves  anomaly semantics fully, leading to a significant divergence between their outputs in anomalous regions. Simultaneously, normal regions exhibit minimal divergence due to consistent reconstruction. Therefore, the Restoration-Identity Discrepancy (RID) is well-suited for anomaly scoring.

\textbf{2. Teacher-Restoration Discrepancy.} When examining the data flow from the teacher encoder to Restoration Decoder in isolation, it resembles a denoising task. Therefore, the Teacher-Restoration Discrepancy (TRD) also exhibits significant discrimination between normal and abnormal images, following DeSTSeg~\cite{destseg}. 


\textbf{3. Anomaly Scoring via Multi-Source Fusion.} For a query image, we firstly obtain a set of initial distance maps from Restoration-Identity or Teacher-Restoration feature pairs, where the value in a map ${M}_{i}$ reflects the point-wise anomaly of the ${i}^{th}$ feature maps. To localize anomalies in a query image,we up-sample ${M}_{RI}^{i}$ or  ${M}_{TR}^{i}$ to image size. Let $\Psi$ denotes the bilinear up-sampling operation. Then the anomaly map RID ${S}_{RI}$ and TRD $S_{TR}$ are formulated as the pixel-wise accumulation of all initial distance maps:
\begin{equation}
{S}_{RI} = \mathop{\sum }\limits_{{i = 1}}^{3}\Psi \left( {M}_{RI}^{i}\right), \quad {S}_{TR} = \sum_{i=1}^3 \Psi\left(M_{TR}^i\right).
\end{equation}

Since both RID and TRD demonstrate effective discriminative capability between normal and anomalous inputs, we directly combine \( S_{RI} \) and \( S_{TR} \) via linear fusion as: $S = \alpha \cdot S_{RI} + (1 - \alpha) \cdot S_{TR}$, where \( \alpha \) is a hyper-parameter controlling the fusion ratio. In addition, the image-level anomaly score is defined as the maximum score of the pixel-level anomaly map $S$.


\begin{table*}[t]
\centering
\small
\caption{Results (\%) on four benchmarking datasets for MUAD. \normalfont \textbf{Bold} and \underline{underline} represent the best and second-best results.}

\label{tab:main_results}
\begin{tabular}{@{}lclccccccccc@{}}
\toprule
\multirow{2}{*}{\textbf{Dataset}} & \multirow{2}{*}{\textbf{Categories}}  & \multirow{2}{*}{\textbf{Method}} & \multicolumn{3}{c}{\textbf{Image-level}} & \multicolumn{5}{c}{\textbf{Pixel-level}} & \multirow{2}{*}{\textbf{Avg}} \\
\cmidrule(lr){4-6} \cmidrule(lr){7-11}
 & & & m-AUROC & mAP & mF1\textsubscript{max} & m-AUROC & mAP & mF1\textsubscript{max} & mAU-PRO & mIoU\textsubscript{max} & \\
\midrule
\multirow{8}{*}{MVTec AD} 
 & \multirow{8}{*}{15}
 & UniAD\textsuperscript{NeurIPS22} & 92.5 & 97.3 & 95.4 & 95.8 & 42.7 & 48.0 & 89.3 & 32.5 & 74.2 \\
 & & DesTSeg\textsuperscript{CVPR23} & 96.4 & 98.6 & 96.2 & 92.0 & \textbf{71.1} & \textbf{68.2} & 83.4 & \textbf{52.8} & \underline{82.3} \\
 & & DiAD\textsuperscript{AAAI24} & 88.9 & 95.8 & 93.5 & 89.3 & 27.0 & 32.5 & 63.9 & 21.1 & 64.0 \\
 & & MambaAD\textsuperscript{NeurIPS24} & 97.8 & \underline{99.3} & \underline{97.3} & 97.4 & 55.1 & 57.6 & \underline{93.4} & 41.2 & 79.9 \\
 & & CNC\textsuperscript{AAAI25} & \textbf{98.6} & \underline{99.3} & - & \underline{98.0} & 56.4 & - & 93.0 & - & - \\
 & & URD\textsuperscript{AAAI25} & 92.2 & 95.8 & 95.9 & 96.8 & 53.0 & 55.1 & 92.6 & 37.9 & 77.4 \\
 & & ViTAD\textsuperscript{CVIU25} & 98.0 & 99.1 & 96.9 & 97.7 & 55.4 & 58.6 & 91.4 & 42.5 & 80.0 \\
 & & MDD-Net (Ours) & \textbf{98.6} & \textbf{99.5} & \textbf{98.0} & \textbf{98.3} & \underline{66.4} & \underline{64.3} & \textbf{93.6} & \underline{48.8} & \textbf{83.4} \\
\midrule
\multirow{8}{*}{VisA} 
 & \multirow{8}{*}{12}
 & UniAD\textsuperscript{NeurIPS22} & 89.0 & 91.0 & 85.8 & 98.3 & 34.5 & 39.6 & 86.5 & 26.4 & 68.9 \\
 & & DesTSeg\textsuperscript{CVPR23} & 89.9 & 91.4 & 86.7 & 86.7 & \textbf{46.6} & \textbf{47.2} & 61.1 & \textbf{32.7} & 67.8 \\
 & & DiAD\textsuperscript{AAAI24} & 84.8 & 88.5 & 86.9 & 82.5 & 17.9 & 23.2 & 44.5 & 14.9 & 55.4 \\
 & & MambaAD\textsuperscript{NeurIPS24} & \textbf{94.5} & \textbf{94.9} & \textbf{90.2} & \underline{98.4} & 39.3 & 43.7 & \textbf{92.1} & 29.5 & \textbf{72.8} \\
 & & CNC\textsuperscript{AAAI25} & \underline{93.2} & 92.6 & - & \textbf{98.5} & 37.8 & - & 91.4 & - & - \\
 & & URD\textsuperscript{AAAI25} & 92.9 & 91.9 & \underline{89.6} & \underline{98.4} & 36.9 & 41.0 & \underline{92.0} & 26.5 & 71.2 \\
 & & ViTAD\textsuperscript{CVIU25} & 90.6 & 91.7 & 86.2 & 98.2 & 36.8 & 41.0 & 84.2 & 27.6 & 69.5 \\
 & & MDD-Net (Ours) & 92.6 & \underline{93.5} & \underline{89.6} & 98.0 & \underline{44.6} & \underline{45.9} & 82.4 & \underline{32.3} & \underline{72.4} \\
\midrule
\multirow{7}{*}{Real-IAD} 
 & \multirow{7}{*}{30}
 & UniAD\textsuperscript{NeurIPS22} & 83.1 & 81.2 & 74.5 & 97.4 & 23.3 & 30.9 & 87.1 & 18.6 & 62.0 \\
 & & DesTSeg\textsuperscript{CVPR23} & 79.3 & 76.7 & 70.7 & 80.3 & \textbf{36.9} & \textbf{40.3} & 56.1 & \textbf{26.2} & 58.3 \\
 & & DiAD\textsuperscript{AAAI24} & 75.6 & 66.4 & 69.9 & 88.0 & 2.9 & 7.1 & 58.1 & 3.7 & 46.5 \\
 & & MambaAD\textsuperscript{NeurIPS24} & \textbf{87.0} & \textbf{85.3} & \textbf{77.6} & \textbf{98.6} & \underline{32.4} & \underline{38.1} & \textbf{91.2} & \underline{23.9} & \textbf{66.8} \\
 & & URD\textsuperscript{AAAI25} & 78.8 & 74.8 & 71.7 & 97.2 & 23.9 & 31.4 & \underline{88.3} & 18.5 & 60.6 \\
 & & ViTAD\textsuperscript{CVIU25} & 82.7 & 80.1 & 73.7 & 97.1 & 24.1 & 32.2 & 84.8 & 19.6 & 61.8 \\
 & & MDD-Net (Ours) & \underline{86.7} & \underline{85.1} & \textbf{77.6} & \underline{97.7} & 30.8 & 35.9 & 86.5 & 22.3 & \underline{65.3} \\
\midrule
\multirow{6}{*}{Uni-Medical} 
 & \multirow{6}{*}{3}
 & UniAD\textsuperscript{NeurIPS22} & 79.0 & 76.1 & 77.1 & 96.6 & 39.3 & 41.1 & 86.0 & 27.6 & 65.4 \\
 & & DesTSeg\textsuperscript{CVPR23} & 78.5 & 77.0 & 78.2 & 65.7 & 41.7 & 34.0 & 35.3 & 21.2 & 54.0 \\
 & & DiAD\textsuperscript{AAAI24} & 78.8 & 77.2 & 77.7 & 95.8 & 34.2 & 35.5 & 84.3 & 23.2 & 63.3 \\
 & & MambaAD\textsuperscript{NeurIPS24} & \underline{83.9} & 80.8 & \underline{81.9} & 96.8 & 45.8 & 47.5 & \underline{88.2} & 33.5 & \underline{69.8} \\
 & & URD\textsuperscript{AAAI25} & 82.8 & \underline{84.1} & 79.6 & 96.4 & 32.7 & 36.9 & 87.7 & 18.8 & 64.9 \\
 & & ViTAD\textsuperscript{CVIU25} & 81.8 & 80.7 & 80.0 & \underline{97.1} & \underline{48.3} & \underline{48.2} & 86.7 & \underline{33.7} & 69.6 \\
 & & MDD-Net (Ours) & \textbf{87.6} & \textbf{87.8} & \textbf{82.0} & \textbf{97.8} & \textbf{60.5} & \textbf{57.8} & \textbf{88.8} & \textbf{41.7} & \textbf{75.5} \\
\bottomrule
\end{tabular}
\end{table*}

\section{Experiments}
\subsection{Experimental Settings}
\textbf{Datasets.} We evaluate our method on four widely used benchmarks for unsupervised anomaly detection: MVTec AD~\cite{mvtec}, VisA~\cite{spot-the-diff}, Real-IAD~\cite{realiad}, and the Uni-Medical benchmark~\cite{bmad,vitad}. MVTec AD~\cite{mvtec} contains 15 objects (5 texture classes and 10 object classes) with a total of 3,629 normal images as the training set and 1,725 images as the test set (467 normal vs. 1,258 anomalous). VisA~\cite{spot-the-diff} contains 12 objects including 8,659 normal training images and 2,162 test images (962 normal vs. 1,200 anomalous). Real-IAD~\cite{realiad} is a large UAD dataset containing 30 distinct objects with 36,465 normal images in the training set and 114,585 images in the test set  (63,256 normal vs. 51,329 anomalous). Uni-Medical~\cite{bmad,vitad} contains three medical imaging modalities: Brain MRI (7,500 normal training images; 640 normal vs. 3,075 anomalous test images), Liver CT (1,542 normal training images; 660 normal vs. 1,258 anomalous test images), and Retinal OCT (4,297 normal training images; 1,041 normal vs. 764 anomalous test images). All datasets provide pixel-level anomaly ground truth masks for evaluation.

\textbf{Metrics.} Following established protocols~\cite{vitad,ader}, we adopt eight evaluation metrics. At the image level, we report mean Area Under the Receiver Operating Curve (mAU-ROC), Average Precision (mAP), and F1 score under the optimal threshold (mF1-max). At the pixel level, we evaluate using mAU-ROC, mAP, mF1-max, the Area Under the Per-Region-Overlap (mAU-PRO) and the order-independent maximal Intersection over Union (mIoU-max) as introduced in InvAD~\cite{invad}. For each dataset, reported scores represent the average performance across all categories.

\textbf{Implementation Details.} We adopt the ViT architecture with DINO-S pretrained weights for the frozen encoder. All experiments are conducted on 256×256 pixel images without other data augmentations except synthetic anomaly augmentation. The AdamW optimizer is employed with an initial learning rate of $1\mathrm{e}^{-4}$, weight decay of $1\mathrm{e}^{-4}$. Our model only requires 100 training epochs on a single GPU in all experiments, and the learning rate drops by 0.1 after 80 epochs. Divergences across datasets exclusively manifest in memory bank capacity and batch dimension configurations: MVTec AD operates with 500 memory slots (batch size 8), both VisA and Uni-Medical implement 1000 memory slots (batch size 4), whereas Real-IAD utilizes 2000 memory slots (batch size 4). The hard shrinkage threshold is uniformly set as 1/N for all datasets, where N corresponds to the memory bank size. Furthermore, the optimal fusion ratios for the MVTec AD, VisA, Real-IAD, and Uni-Medical datasets are 0.4, 0.4, 0.1 and 0.5, respectively. To improve training efficiency and convergence, we adopt an online hard example mining strategy by discarding gradient updates for well-reconstructed feature points exhibiting low cosine distances.

\begin{figure*}[t] 
\centering
\includegraphics[width=0.9\textwidth]{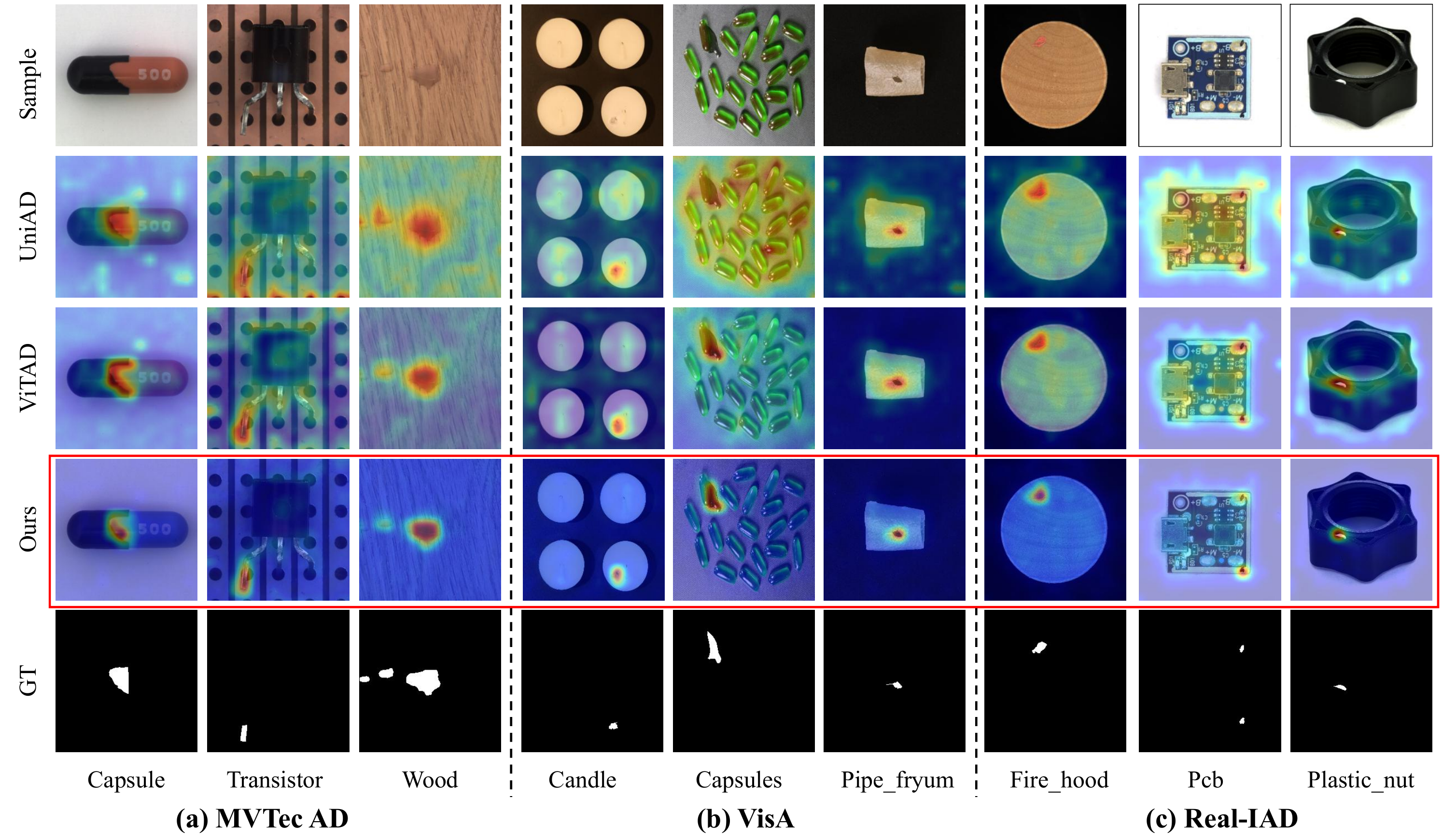} 
\caption{Visualization for detection results of different methods on MVTec AD/VisA/Real-IAD datasets.}
\Description[Visualization for detection results of different methods on MVTec AD/VisA/Real-IAD datasets.]{Visualization for detection results of different methods on MVTec AD/VisA/Real-IAD datasets.}
\label{fig:vis_all}
\end{figure*}

\begin{table*}[t]
\centering
\small
\caption{Results (\%) of Category-Specific Anomaly Detection Performance on Uni-Medical. \normalfont The numerical values in \textcolor{red}{red brackets} denote the performance gain of MDD-Net over ViTAD~\cite{vitad}.}

\label{tab:medical_per_class}
\begin{tabular}{@{}clccccccccc@{}}
\toprule
\multirow{2}{*}{\textbf{Category}} & \multicolumn{3}{c}{\textbf{Image-level}} & \multicolumn{5}{c}{\textbf{Pixel-level}} & \multirow{2}{*}{\textbf{Avg}} \\
\cmidrule(lr){2-4} \cmidrule(lr){5-9}
& m-AUROC & mAP & mF1\textsubscript{max} & m-AUROC & mAP & mF1\textsubscript{max} & mAU-PRO & mIoU\textsubscript{max} & \\
\midrule
Brain 
& 89.52\textcolor{red}{\scriptsize{(+5.64)}} & 97.37\textcolor{red}{\scriptsize{(+1.59)}} & 93.06\textcolor{red}{\scriptsize{(+1.28)}} 
& 97.70\textcolor{red}{\scriptsize{(+0.83)}} & 61.66\textcolor{red}{\scriptsize{(+11.98)}} & 58.79\textcolor{red}{\scriptsize{(+7.76)}} 
& 83.76\textcolor{red}{\scriptsize{(+3.54)}} & 41.63\textcolor{red}{\scriptsize{(+8.24)}} & 77.94\textcolor{red}{\scriptsize{(+5.11)}} \\

Liver 
& 64.05\textcolor{red}{\scriptsize{(+10.23)}} & 53.64\textcolor{red}{\scriptsize{(+19.08)}} & 65.65\textcolor{red}{\scriptsize{(+1.85)}} 
& 97.82\textcolor{red}{\scriptsize{(+0.57)}} & 12.14\textcolor{red}{\scriptsize{(+23.48)}} & 21.05\textcolor{red}{\scriptsize{(+18.32)}} 
& 91.52\textcolor{red}{\scriptsize{(+2.32)}} & 11.76\textcolor{red}{\scriptsize{(+12.75)}} & 52.20\textcolor{red}{\scriptsize{(+11.07)}} \\

Retinal 
& 92.27\textcolor{red}{\scriptsize{(+1.16)}} & 90.24\textcolor{red}{\scriptsize{(+1.52)}} & 82.56\textcolor{red}{\scriptsize{(+1.58)}} 
& 95.68\textcolor{red}{\scriptsize{(+0.70)}} & 69.65\textcolor{red}{\scriptsize{(+2.71)}} & 62.82\textcolor{red}{\scriptsize{(+4.54)}} 
& 81.41\textcolor{red}{\scriptsize{(+3.80)}} & 45.80\textcolor{red}{\scriptsize{(+4.99)}} & 77.55\textcolor{red}{\scriptsize{(+2.63)}} \\

\cmidrule[0.6pt](lr){1-10}
Avg 
& 81.95\textcolor{red}{\scriptsize{(+5.68)}} & 80.42\textcolor{red}{\scriptsize{(+7.40)}} & 80.42\textcolor{red}{\scriptsize{(+1.57)}} 
& 97.07\textcolor{red}{\scriptsize{(+0.70)}} & 47.81\textcolor{red}{\scriptsize{(+12.72)}} & 47.55\textcolor{red}{\scriptsize{(+10.21)}} 
& 85.56\textcolor{red}{\scriptsize{(+3.22)}} & 33.06\textcolor{red}{\scriptsize{(+8.66)}} & 69.23{\textcolor{red}{\scriptsize{(+6.27)}}} \\
\bottomrule
\end{tabular}
\end{table*}

\subsection{Main Results}
We compare our method with seven SoTA reconstruction-based methods, including UniAD~\cite{uniad}, DeSTSeg~\cite{destseg}, DiAD~\cite{diad}, MambaAD~\cite{mambaad}, CNC~\cite{cnc}, URD~\cite{urd} and ViTAD~\cite{vitad}. Table \ref{tab:main_results} reports the results on the four datasets. For the MVTec AD dataset, our MDD-Net outperforms all the comparative methods in multi-class anomaly detection and localization. Specifically, compared to SoTA MambaAD~\cite{mambaad}, our proposed MDD-Net shows an improvement of 0.8 ↑/0.2 ↑/0.7 ↑ at image-level and 0.9 ↑/11.3 ↑/6.7 ↑/0.2 ↑/7.6 ↑at pixel-level. The VisA dataset is more complex and challenging, yet our method still demonstrates excellent performance. Our MDD-Net exceeds the performance of  MambaAD~\cite{mambaad} by 5.3 ↑/2.2 ↑/2.8 ↑ in pixel-level mAP/mF1-max/mIoU-max metrics, demonstrating the superiority in localization accuracy. In addition, MDD-Net achieves performance gain of 4.0 ↑/5.0 ↑/3.9 ↑/0.6 ↑ using image-level metrics on Real-IAD dataset, compared with the baseline ViTAD~\cite{vitad}. This illustrates the scalability, versatility, and effectiveness of our method. Compared to industrial AD datasets, Uni-Medical presents more significant challenges due to the more difficult anomaly types and a more comprehensive range of anomaly areas. And MDD-Net achieves a significant advantage, reaching 75.5 in the mean metric that surpasses the second-best MambaAD~\cite{mambaad} by +5.7 ↑. This indicates that our MDD-Net has strong generalization ability across different types of datasets. Figure \ref{fig:vis_all} shows the anomaly localization results on various categories on the MVTec AD, VisA and Real-IAD datasets. Compared to UniAD~\cite{uniad} and ViTAD~\cite{vitad}, our MDD-Net can find more accurate and compact anomalous areas with less edge uncertainty and shows fewer false positives in normal areas. Using examples of a textural wood with background interference and a capsule object with irregularly shaped anomalies, MDD-Net segments anonymous areas more accurately with fewer false positive responses.

\subsection{More Detailed Results on Uni-Medical}
\begin{figure}[t] 
  \centering
  \includegraphics[width=0.9\columnwidth]{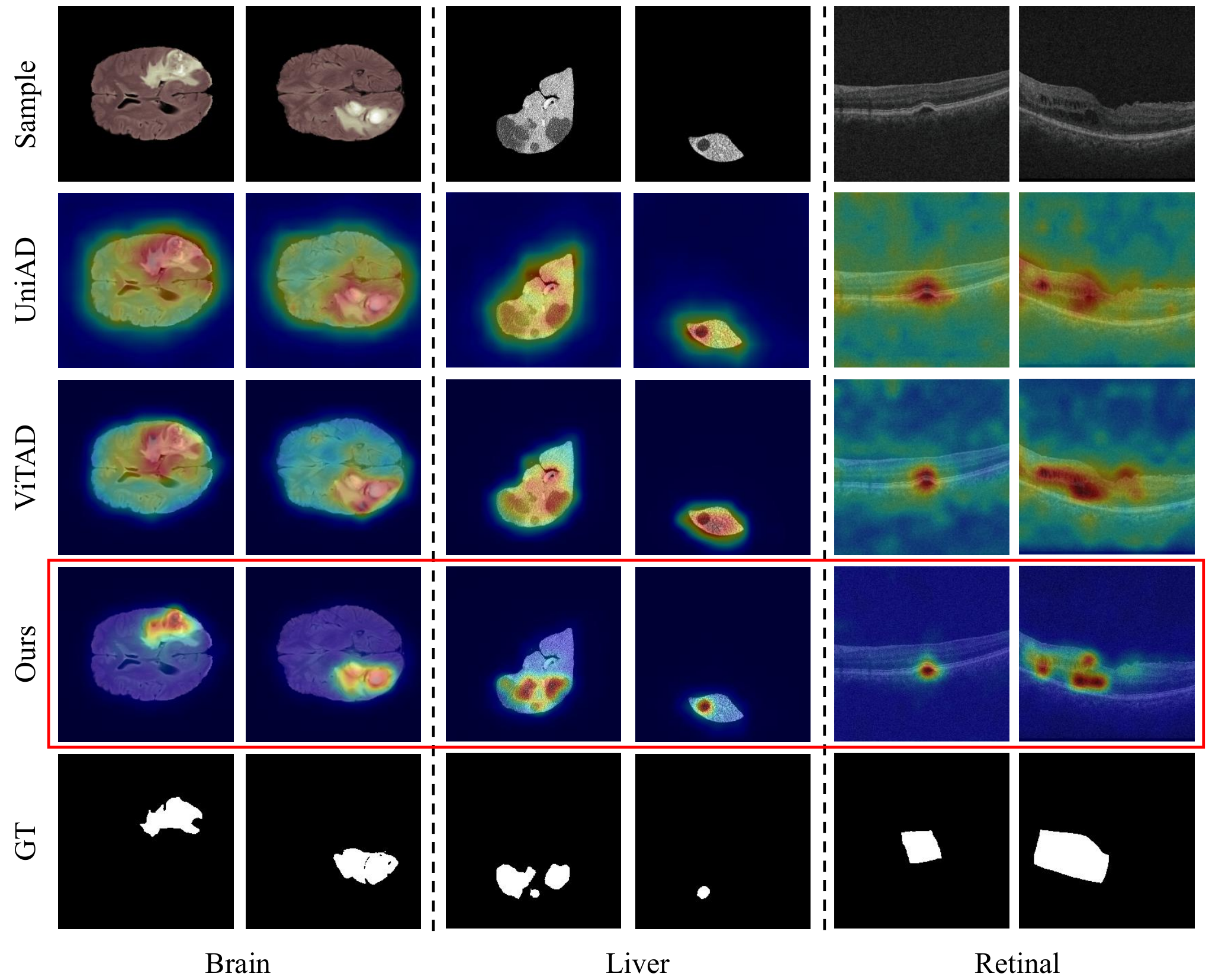} 
  \caption{Visualization for detection results of different methods on Uni-Medical.}
  \Description[Visualization for detection results of different methods on Uni-Medical.]{Visualization for detection results of different methods on Uni-Medical.}
  \label{fig:vis_unimedical}
\end{figure}

Table \ref{tab:medical_per_class} provides a comparison between the baseline ViTAD~\cite{vitad} and our proposed MDD-Net using both image-level and pixel-level metrics across three anatomical categories (brain, liver, retinal) on the Uni-Medical dataset.  Our method exhibits significant enhancements across all metrics for each category, particularly in pixel-level measurements. Specifically, MDD-Net achieves improvements of +12.72 ↑/+10.21 ↑/+8.66 ↑ in pixel-level mAP/mF1-max/mIoU-max metrics compared to ViTAD~\cite{vitad}. These comprehensive improvements confirm the excellent generalization capacity of MDD-Net when handling heterogeneous medical images. The visualization results in  Figure~\ref{fig:vis_unimedical} show that our method achieves more precise anomaly segmentation boundaries with significantly reduced localization errors. This accurate localization capability enables our method to generate reliable annotations for semantic segmentation tasks through weak supervision paradigms~\cite{mseg,mseg1,mseg2}, which is valuable for few-shot medical image analysis~\cite{me_anls,me_anls2}.

\begin{table}[t]
\centering
\caption{Overall Ablation Results(\%) on MVTec AD }
\label{tab:ablation_all}
\begin{threeparttable}
\footnotesize
\begin{tabular}{@{}ccc@{\ }cccc@{}}
\toprule
\textbf{CMM} & \textbf{DRD-Net-L} & \textbf{DRD-Net+L} & I-\textbf{mAUROC} & \textbf{P-mAP} & \textbf{P-mIoU} & \textbf{Avg} \\
\midrule
 \textendash & \textendash & \textendash & 98.03 & 55.43 & 42.52 & \textbf{65.33} \\
 $\checkmark$ & \textendash & \textendash & 98.85 & 60.06 & 45.41 & \textbf{68.11} \\
 \textendash & $\checkmark$ & \textendash & 98.12 & 56.73 & 43.64 & \textbf{66.16} \\
 \textendash & \textendash & $\checkmark$ & 97.34 & 61.91 & 44.90 & \textbf{68.05} \\
 $\checkmark$ & \textendash & $\checkmark$ & 98.62 & 66.40 & 48.76 & \textbf{71.26} \\
\bottomrule
\end{tabular}
\begin{tablenotes}
\item[†] DRD-Net$\pm$L denote the exclusion/inclusion of the $\mathbf{L}_{dist}$ respectively.
\end{tablenotes}
\end{threeparttable}
\end{table}

\begin{table}[t]
\small
\centering
\caption{Memory Updating Results (\%) on MVTec AD}
\label{tab:ab_mem}
\begin{tabularx}{\columnwidth}{Ycccc}
\toprule
\textbf{MemoryUpdate} & \textbf{{I-mAUROC}} & \textbf{{P-mAP}} & \textbf{{P-mIoU}} & \textbf{{Avg}} \\
\midrule
\textbf{CMM\_N2I} & 98.36  & 66.32  & 48.36  & \textbf{71.01} \\
\textbf{Mem\_N2R} & 98.55  & 66.12  & 48.50  & \textbf{71.05} \\
\textbf{CMM\_N2R} & 98.62  & 66.40  & 48.76  & \textbf{71.26} \\
\bottomrule
\end{tabularx}
\end{table}

\subsection{Ablation Studies}
We conduct systematic experiments to evaluate the effectiveness of the proposed components, including the Dual-Decoder Reverse Distillation Network (DRD-Net$\pm$L,  with/without $\mathcal{L}_{dist}$  loss) and the Class-aware Memory Module (CMM), with ViTAD~\cite{vitad} as the baseline (first row in the table). The results on MVTec AD are reported in Table~\ref{tab:ablation_all}, from which we can see that both CMM and DRD-Net variants (DRD-Net$\pm$L) can directly contribute to the model performance. Notably, the significant improvements of P-mAP and P-mIoU prove the localization superiority of the dual branch over the single branch. The complete MDD-Net configuration achieves 66.40\% pixel-level mAP and 48.76\% mIoU on MVTec AD dataset.

\begin{figure}[t] 
  \centering
  \includegraphics[width=\columnwidth]{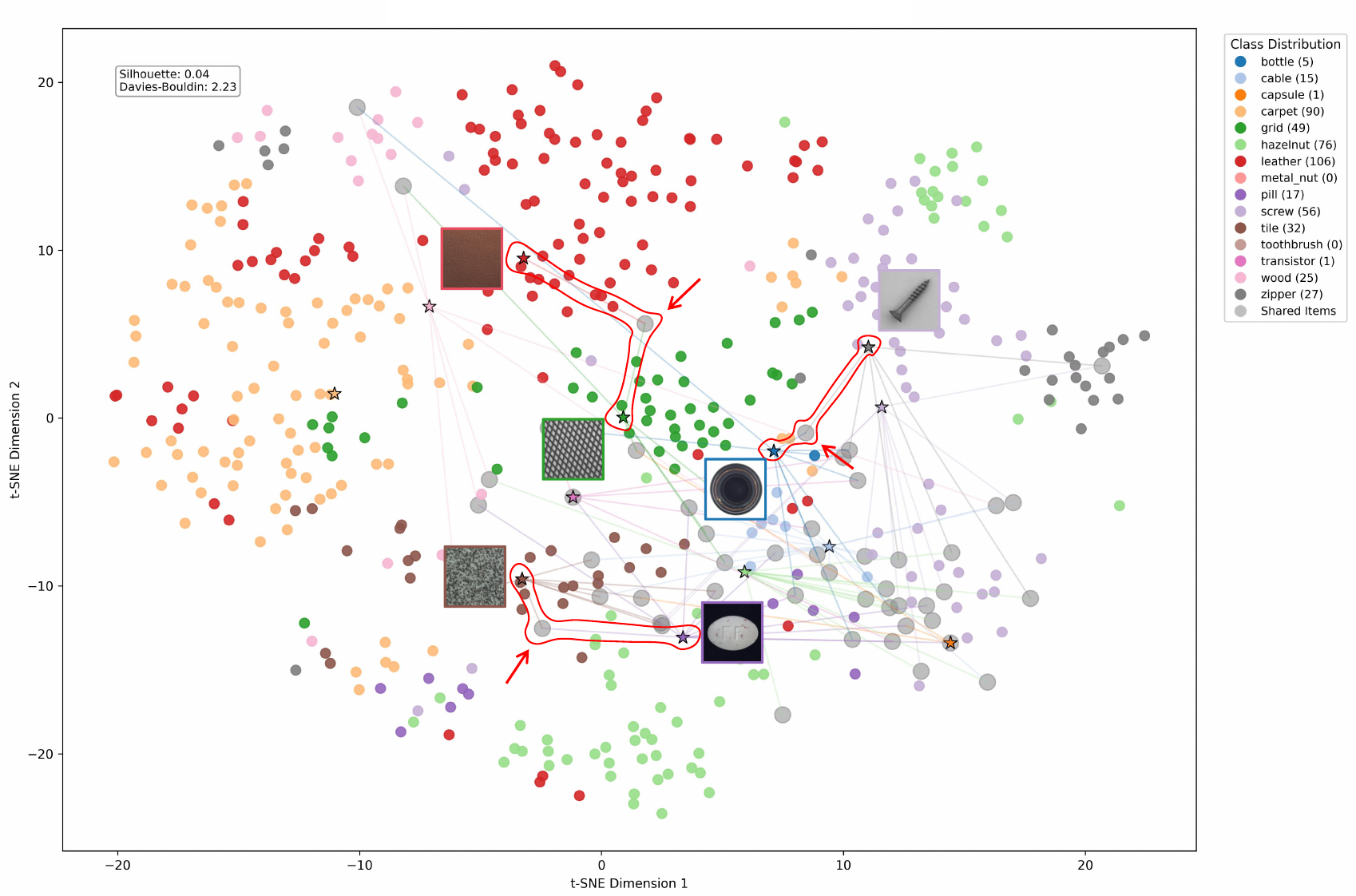} 
  \caption{Visualization of class-specific and class-shared memory items of CMM. Memory items are categorized by entropy thresholds (top 10\% entropy as class-shared gray dots, remainder as class-specific colored dots). Class-specific items are assigned to their maximum-probability classes (color-coded), while shared items connect to their two most probable class prototypes (solid lines), indicating inter-class knowledge. Red arrows highlight class-shared patterns, e.g., leather$\leftrightarrow$grid. Cluster validity is quantified by silhouette score~\cite{cluster1} and Davies-Bouldin index~\cite{clsuter2}.}
  \Description[Clustering]{Clustering}
  \label{fig:cmm}
\end{figure}

To validate the effectiveness of our proposed Class-aware Memory Module (CMM), we evaluate three distinct configurations and update mechanisms: 1) \textbf{CMM\_N2I}: employing the CMM where Normal samples are fed into the Identity Decoder for reconstruction. 2) \textbf{Mem\_N2R}: utilizing Conventional Memory (Mem) where Normal samples are fed into the Restoration Decoder for reconstruction. 3) \textbf{CMM\_N2R}: employing the CMM where Normal samples are fed into the Restoration Decoder for reconstruction. As shown in Table \ref{tab:ab_mem}, \textbf{CMM\_N2R} achieves superior performance, outperforming other variants by 0.21-0.25\% on average. Furthermore, we conducted cluster analysis of memory items in CMM using t-SNE (as illustrated in Figure \ref{fig:cmm}). The visualization results demonstrate that the learned memory prototypes effectively balance intra-class specificity with inter-class commonality preservation. More visualizations are available in the Supplementary Materials Section B.2. 


\section{Conclusion}
In this work, we proposed MDD-Net, a novel memory-augmented dual-decoder framework designed for multi-class unsupervised anomaly detection. By integrating the Dual-Decoder Reverse Distillation Network (DRD-Net) and the Class-aware Memory Module (CMM), our method effectively addresses two persistent challenges in reconstruction-based UAD: the insufficient reconstruction of hard-normal features and the over-generalization of subtle anomalies. Through comprehensive experiments across four diverse benchmarks and eight evaluation metrics, MDD-Net demonstrates consistent improvements over state-of-the-art methods in both anomaly detection and localization.


\textbf{Limitations and Future Work.} Despite the strong empirical results, our approach inherits a key limitation from HVQ-Trans \cite{hvq-trans}—the reliance on category labels during training. Future work should explore integrating clustering mechanisms to eliminate dependency on labeled categories. Additionally, the current model’s computational efficiency remains suboptimal for real-world deployment, highlighting the need for lightweight architectures.

\end{document}


\title{Supplementary Materials of Memory-Augmented Dual-Decoder Networks for Multi-Class Unsupervised Anomaly Detection}

\maketitle

\appendix
This supplementary material provides additional descriptions of the proposed MDD-Net, including implementation details and empirical results.

\section{Additional implementation details}
\label{sec:impl_details}
\subsection{Normality Prototypes Learning of CMM}
\ 

\begin{figure}[ht] 
  \centering
  \includegraphics[width=\columnwidth]{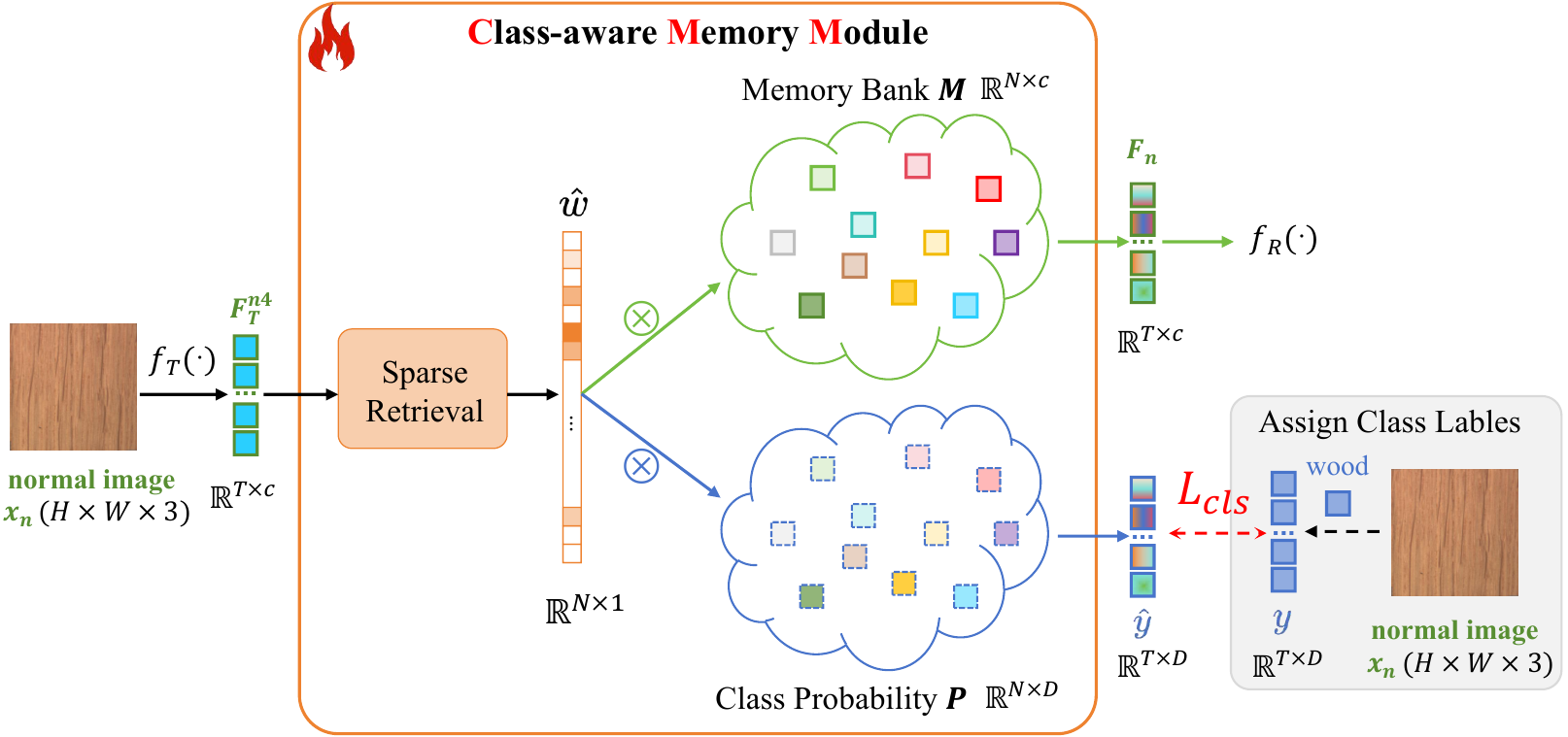} 
  \caption{In CMM, we use global labels to provide token-level supervision, disentangling category-specific feature patterns.}
  \Description[The proposed CMM.]{Classification loss of the proposed CMM.}
  \label{fig:cmmloss}
\end{figure}

The workflow of the classification loss in \textit{Normality Prototypes Learning} of the CMM is illustrated in Figure \ref{fig:cmmloss}.

\subsection{Inferring Process of MDD-Net.}
\ 

\begin{figure*}[htbp] 
\centering
\includegraphics[width=0.9\textwidth]{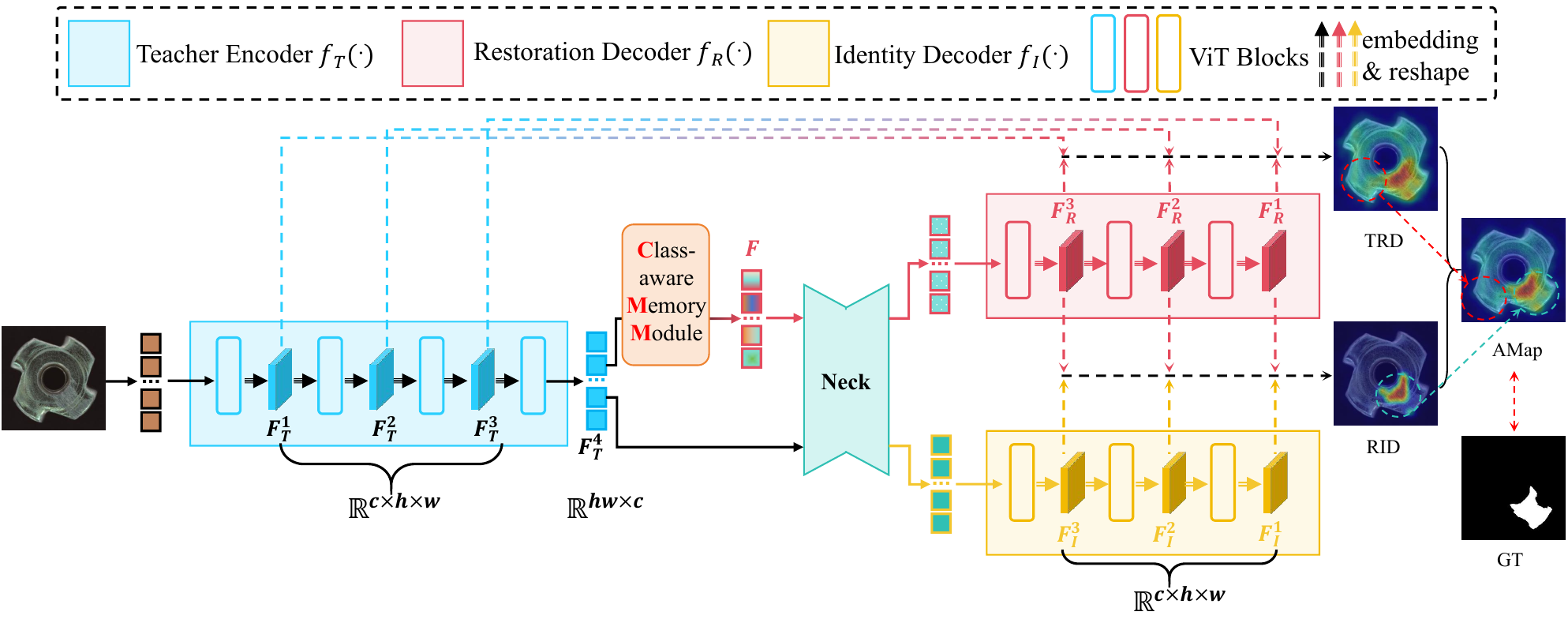} 
\caption{Inferring Process of MDD-Net. Initially, a test image is fed into the Teacher Encoder to extract multi-stage feature maps. These extracted features are then routed to two distinct branches: "CMM$\rightarrow$Neck$\rightarrow$Restoration Decoder" and "Neck$\rightarrow$Identity Decoder", which derive hierarchical feature maps, respectively.
Ultimately, the Restoration-Identity Discrepancy (RID) and Teacher-Restoration Discrepancy (TRD) are fused to generate the final anomaly map.}
\label{fig:infer}
\end{figure*}

During inference, the test image is first processed by the Teacher Encoder to extract multi-stage feature maps. These extracted features are subsequently routed to the Restoration Decoder and Identity Decoder to derive hierarchical feature maps, respectively. Ultimately, we compute two complementary discrepancies: the Restoration-Identity Discrepancy (RID) and Teacher-Restoration Discrepancy (TRD), which are fused to generate the final anomaly map. The complete workflow is illustrated in Figure \ref{fig:infer}.

\begin{figure}[htbp] 
\centering
\includegraphics[width=0.8\columnwidth]{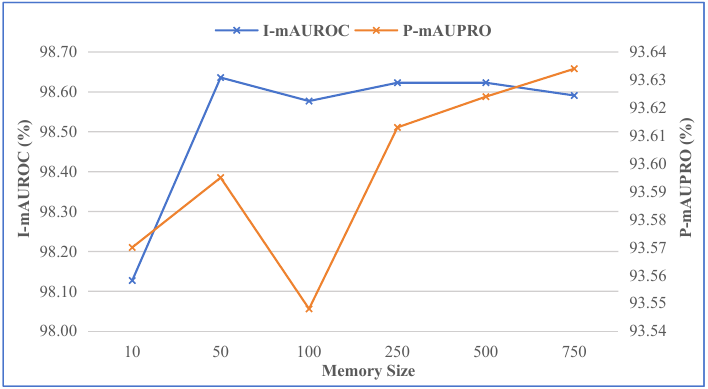} 
\caption{AUROC \& AUPRO scores of MDD-Net with different memory size on MVTec AD.}
\label{fig:memsize}
\end{figure}

\subsection{Algorithm Implementation}
\ 

The algorithm of our MDD-Net during the training (Phase 2) and inferring (Phase 3) phases is presented in algorithm\ \ref{alg:cmdvd_algorithm}.

\label{sec:pseudocode}
\begin{algorithm}
\caption{MDD-Net Training and Inference Framework}
\label{alg:cmdvd_algorithm}
\begin{algorithmic}[0]
\REQUIRE Normal image dataset $\mathbb{X}$, Query image $x$
\ENSURE Anomaly localization map $S$, Image-level score $s$

\STATE \textbf{Phase 1: Network Initialization}
\STATE \indentbullet Teacher Encoder: $f_{T}(\cdot)$ (Frozen 12-layer ViT)
\STATE \indentbullet Neck module: $Neck\in \mathbb{R}^{c \times c}$ 
\STATE \indentbullet Memory module: $\mathcal{M} = \{ m_i \}_{i=1}^N$ with $m_i \in \mathbb{R}^c$
\STATE \indentbullet Restoration Decoder: $f_{R}(\cdot)$ (9-layer ViT)
\STATE \indentbullet Identity Decoder:$f_{I}(\cdot)$ (9-layer ViT)

\STATE \textbf{Phase 2: Training Process}
\FOR{$epoch < max\_epochs$}
    \FOR{each $(x_n, y) \in \mathbb{X}$}
        \STATE \textcolor{blue}{\COMMENT{1. Anomaly Generation}}
        \STATE $(x_a, M_{gt}) \gets \mathcal{A}(x_n)$ 
        
        \STATE \textcolor{blue}{\COMMENT{2. Feature Extraction}}
        \STATE $\{F_{T}^{ni}\}_{i=1}^4 \gets f_T(x_n)$ 
        \STATE $\{F_{T}^{ai}\}_{i=1}^4 \gets f_{T}(x_a)$
        
        \STATE \textcolor{blue}{\COMMENT{3. Anomaly Suppression}}
        \STATE $F_{a} \gets \mathcal{M}(F_{T}^{a4})$
        \STATE $F_a \gets Neck(F_a)$
        
        \STATE \textcolor{blue}{\COMMENT{4. Dual-Decoder Distillation}}
        \STATE $\{F_{R}^{ai}\}_{i=1}^3 \gets f_R(F_a)$
        \STATE $\{F_{I}^{ai}\}_{i=1}^3 \gets f_{I}(F_T^{a4})$
        
        \STATE \textcolor{blue}{\COMMENT{5. Loss Computation}}
        \STATE $\mathcal{L}_{restoration} \gets \sum_{i=1}^3 \mathcal{D}_{\cos}(F_R^{ai}, F_T^{ni})$
        \STATE $\mathcal{L}_{identity} \gets \sum_{i=1}^3 \mathcal{D}_{\cos}(F_I^{ai}, F_T^{ai})$
        \STATE $\mathcal{L}_{dist} \gets  \sum_{i=1}^3\| M_{RI}^i - M_{gt}^i \|_1$
        \STATE $\mathcal{L}_{rec} \gets \sum_{i=1}^3 \mathcal{D}_{\cos}(f_R(Neck(\mathcal{M}(F_T^{n4})), F_T^{ni})$
        \STATE $\mathcal{L}_{cls} \gets CrossEntropy(\hat{y}, y)$
        \STATE \textcolor{blue}{\COMMENT{6. Parameter Update}}
        \STATE $\nabla \left[ \sum_{k\in\{restoration,identity,dist,rec,cls\}}  \mathcal{L}_k \right]$
    \ENDFOR
\ENDFOR

\STATE \textbf{Phase 3: Anomaly Detection}
\STATE \textcolor{blue}{\COMMENT{1. Feature Extraction}}
\STATE $\{F_{\mathrm{T}}^{i}\}_{i=1}^4 \gets f_{\mathrm{T}}(x)$
\STATE \textcolor{blue}{\COMMENT{2. Dual-Decoder Distillation}}
\STATE $\{F_R^{i}\}_{i=1}^3 \gets f_R(Neck(\mathcal{M}(F_T^4)))$ 
\STATE $\{F_I^{i}\}_{i=1}^3 \gets f_I(Neck(F_T^4))$ 
\STATE \textcolor{blue}{\COMMENT{3. Anomaly Scoring}}
\FOR{$i = 1$ to $3$} 
    \STATE $M_{RI}^i \gets \mathcal{D}_{\cos}(F_R^{i}, F_I^{i})$, $M_{TR}^i \gets \mathcal{D}_{\cos}(F_T^{i}, F_R^{i})$ 
\ENDFOR
\STATE $S_{RI} \gets \sum_{i=1}^3 M_{RI}^i$, $S_{TR} \gets \sum_{i=1}^3 M_{TR}^i$

\STATE $S \gets \alpha \cdot S_{RI} + (1-\alpha) \cdot S_{TR}$, $s = max(S)$

\end{algorithmic}
\end{algorithm}

\section{Additional experimental results}
\label{sec:additional_results}

\subsection{Robustness to the number of prototypes}
\label{exp:mem_size}
\ 

We use the MVTec AD dataset to study the robustness of the proposed MDD-Net to the memory size N . We conduct the experiments by using different memory size settings and show the image-level mAUROC values and the pixel-level mAUPRO values in Figure \ref{fig:memsize} . Given a large enough memory size, the MDD-Net can robustly produce plausible results.


\subsection{CMM Visualizations}
\label{sec:cmm_share_vis}
\

\begin{figure}[htbp]
\centering
\subcaptionbox{carpet\label{fig:ClassShare1}}[0.45\columnwidth]{
  \includegraphics[width=\linewidth]{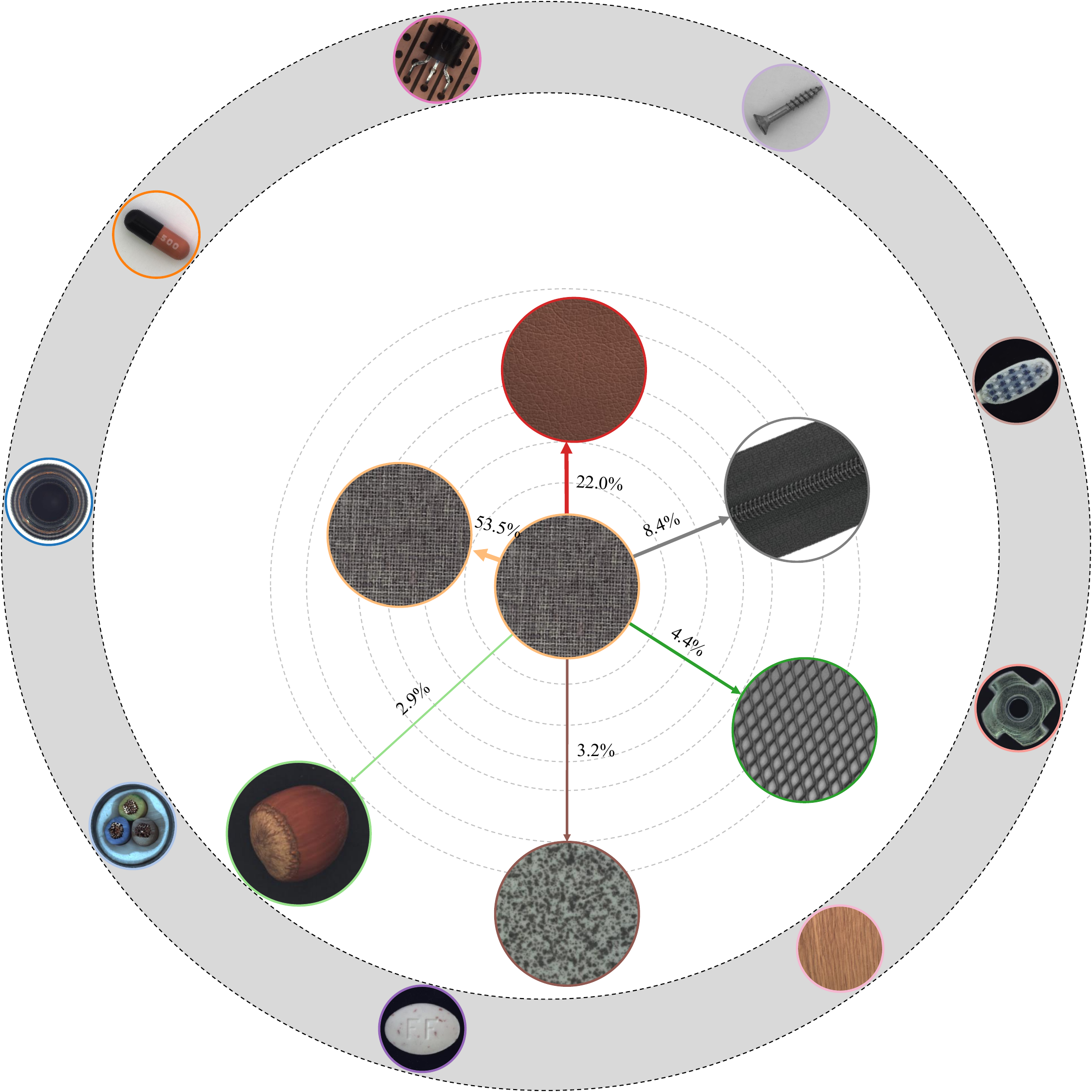}
}
\hfill
\subcaptionbox{hazelnut\label{fig:ClassShare2}}[0.45\columnwidth]{
  \includegraphics[width=\linewidth]{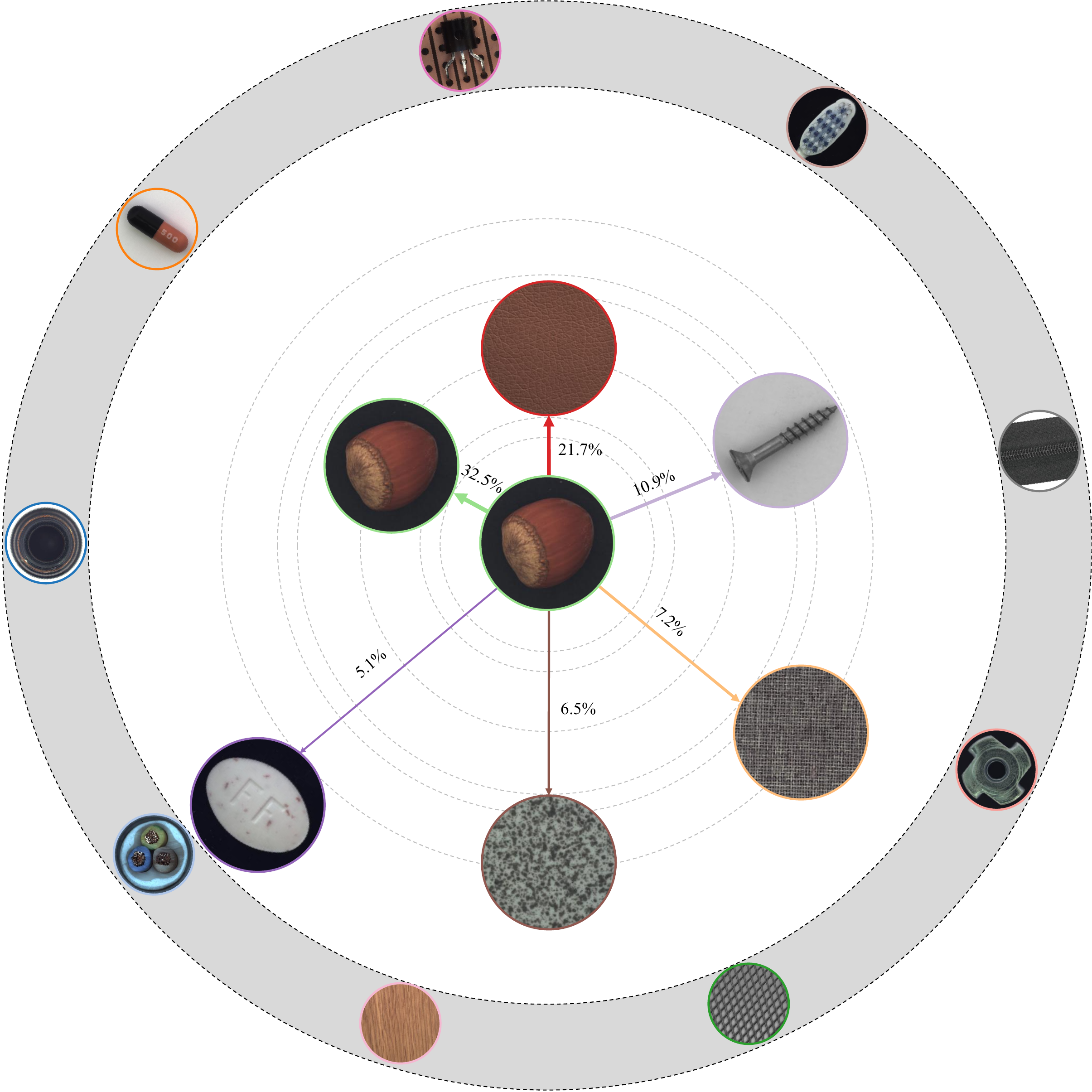}
}
\caption{Cross-Category Feature Sharing in CMM. The CMM dynamically replaces target categories (center) by combining memory prototypes from multiple classes (surrounding) adaptively. Textural carpet samples mainly utilize textural prototypes (carpet, leather, grid, tile) and high-frequency object prototypes (zipper, hazelnut). Conversely, the hazelnut objects leverage both object prototypes (hazelnut, screw, pill) and textural ones (leather, carpet), thereby suggesting shared inter-class characteristics among these categories. Arrow weights and spatial proximity quantify prototype utilization frequency and inter-class feature-sharing intensity, respectively.}
\label{fig:CMMShare}
\end{figure}

To further demonstrate the cross-class shared features in CMM, we statistically analyzed and visualized the averaged utilization patterns of memory items during the reconstruction for the (texture) carpet and (object) hazelnut, as shown in figure \ref{fig:CMMShare}.

